\documentclass[runningheads]{llncs}
\usepackage{graphicx}
\usepackage{comment}
\usepackage{amsmath,amssymb} 
\usepackage{color}

\usepackage[width=122mm,left=12mm,paperwidth=146mm,height=193mm,top=12mm,paperheight=217mm]{geometry}
\usepackage[pagebackref=true,breaklinks=true,letterpaper=true,colorlinks,bookmarks=false]{hyperref}
\usepackage{multirow}

\begin{document}
\pagestyle{headings}
\mainmatter
\def\ECCVSubNumber{1462}  

\title{PBRnet: Pyramidal Bounding Box Refinement to Improve Object Localization Accuracy}

\titlerunning{ECCV-20 submission ID \ECCVSubNumber}
\authorrunning{ECCV-20 submission ID \ECCVSubNumber}
\author{Anonymous ECCV submission}
\institute{Paper ID \ECCVSubNumber}

\titlerunning{Abbreviated paper title}
%
\author{Li Xiao\inst{1,2} \thanks{equal contribution}\and Yufan Luo \inst{1,2}\textsuperscript{$\star$} \and Chunlong Luo \inst{1,2} \and Lianhe Zhao \inst{1,2}, Quanshui Fu \inst{3}, Guoqing Yang \inst{3}, Anpeng Huang \inst{4}, Yi Zhao \inst{1,2}}
\authorrunning{L. Xiao et al.}
%
\institute{Key Laboratory of Intelligent Information Processing, Advanced Computer Research Center, Institute of Computing Technology, Chinese Academy of Sciences, Beijing, China \\
\email{xiaoli@ict.ac.cn}
 \and
School of Computer and Control Engineering, University of Chinese Academy of Sciences (UCAS), Beijing, China \\
 \and
Department of Radiology, Suining Central Hospital \and 
Peking University
}
\maketitle

\begin{abstract}
Many recently developed object detectors focused on coarse-to-fine framework which contains several stages that classify and regress proposals from coarse-grain to fine-grain, and obtains more accurate detection gradually.  Multi-resolution models such as Feature Pyramid Network(FPN) integrate information of different levels of resolution and effectively improve the performance.  Previous researches also have revealed that localization can be further improved by: 1) using fine-grained information which is more translational variant; 2) refining local areas which is more focused on local boundary information. Based on these principles, we designed a novel boundary refinement architecture to improve localization accuracy by combining coarse-to-fine framework with feature pyramid structure, named as Pyramidal Bounding Box Refinement network(PBRnet), which parameterizes gradually focused boundary areas of objects and leverages lower-level feature maps to extract finer local information when refining the predicted bounding boxes. Extensive experiments  are performed on the MS-COCO dataset. The PBRnet brings a significant performance gains by roughly 3 point of $mAP$ when added to FPN or Libra R-CNN. Moreover, by treating Cascade R-CNN as a coarse-to-fine detector and replacing its localization branch by the regressor of PBRnet, it leads an extra performance improvement by 1.5 $mAP$, yielding a total performance boosting by as high as 5 point of $mAP$.  Code will be made available.
\keywords{Coarse-to-Fine Detection,Multi-Resolution Localization, Pyramidal Bounding Box Refinement}
\end{abstract}

\section{Introduction}
Object detection\cite{deng2009imagenet,lin2014microsoft,everingham2010pascal} serves as a fundamental task with many applications in computer vision, and has been rapidly improved in recent years based on the deep learning models\cite{peng2018megdet,krizhevsky2012imagenet,simonyan2014very,szegedy2015going,he2016deep,szegedy2017inception,xie2017aggregated}. Among them Faster R-CNN is a classical two stage detection framework which has been most widely used and is the basis of many advanced models\cite{girshick2015fast,girshick2014rich,ren2015faster,singh2018analysis,dai2016r,gidaris2015object,he2017mask,shrivastava2016training,lin2017focal,dai2017deformable}.  It formulates the detection as two procedures:1)region proposal network produces high quality proposals by distinguishing anchors between foreground and background as well as regressing them to the assigned ground truth locations;2)anchor generated Region-of-Interests(RoIs) are fine classified and their coordinates are further refined by a constraint such as cross-entropy or Smooth L1 loss.

Several researches have developed the coarse-to-fine framework to improve the performance of object detection models. Iterative bounding box regression\cite{gidaris2016attend} claims that a single workflow is not sufficient for detection and iteratively applies additional post-processing steps. Cascade R-CNN\cite{cai2018cascade}, which is served as one of the best single model detector, designs a multi-stage architecture by iteratively increasing the IoU threshold to train the detector(named as quality of the detector). Recently, Side-Aware Boundary Localization (SABL)\cite{wang2019side} demonstrates that local areas are more suitable for boundary refinement, and proposes a lightweight two step bucketing scheme for localization which searches the bucket where the boundary resides and then predicts the offsets of the bucket.

Feature Pyramid Network(FPN)\cite{lin2017feature,pang2019libra,qin2019thundernet,liu2018path} is an effective backbone which uses a top-down architecture with lateral connections to build an in-network feature pyramid structure. It surpasses other detectors a large margin by using a multi-solution detector and assigning each proposal to an appropriate map according to its scale. The Faster R-CNN achieves excellent gains in both accuracy and speed by using FPN backbone for feature extraction.

It is a fact that feature maps of different levels in hierarchical feature pyramid integrate information from different grains, the diverse information concealed in the pyramid structure provides a way to further improve the object detection performance.On the one hand, lower-level feature map with high resolution consists more information of an object, and thus are capable for extracting finer-grained description compare with higher-level feature maps. On the other hand, as pointed in \cite{cheng2018revisiting}, localization requires translational variant features, the information in the lower-level feature map is more sensitive to displacement and translation as the smaller stride compare with that in the higher-level feature maps. Therefore, the performance of localization could be improved by refining the predicted bounding boxes on more fine-grained feature maps after detecting the objects.

\begin{figure}
\begin{center}
\includegraphics[width=\textwidth]{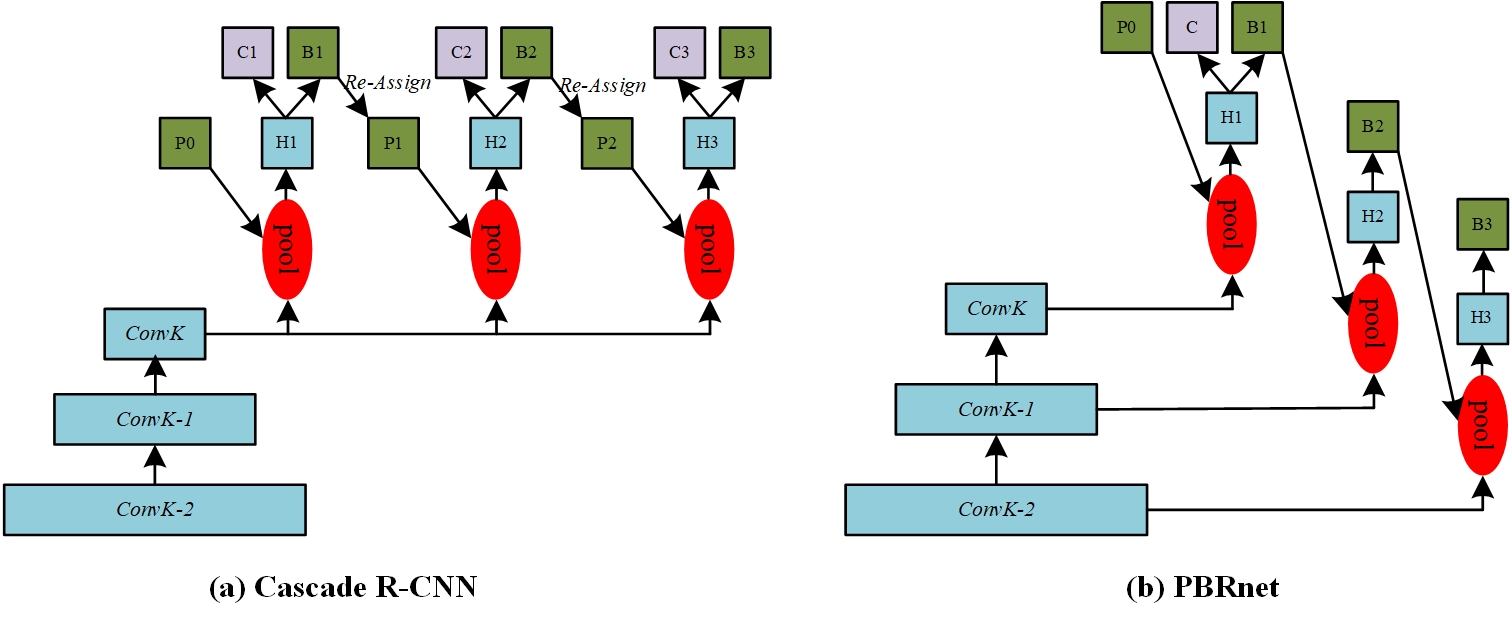}
\end{center}
   \caption{The comparison between architectures of Cascade R-CNN and PBRnet. "$ConvK,ConvK-1,ConvK-2$'' represent feature pyramid, ``pool'' region-wise feature extraction, ``H'' network head, ``B'' bounding box, and ``C'' classification. ``B0'' is proposals in all architectures. ``Re-Assign'' indicates the process of re-assigning ground truth to proposals with larger IoU threshold.}
\label{fig1}
\end{figure}

In this paper, we combined coarse-to-fine framework with multi-resolution detector, and proposed a new localization architecture, Pyramidal Boundary Refinement Network(PBRnet), based on the Faster-R-CNN-FPN framework. It decomposes localization branch into a coarse-to-fine framework that each procedure extracts gradually focused local boundary areas  and leverages feature maps of lower level(same level if it is already on the lowest level) to refine the proposal to its assigned ground truth. Similar to the Cascade R-CNN\cite{cai2018cascade}, PBRnet also iteratively refines the bounding boxes but there are three advantages of the PBRnet over the Cascade R-CNN as illustrated in Figure \ref{fig1}.  Firstly, assuming a proposal is firstly assigned to the $ConvK$th feature map,  different from the Cascade R-CNN in which all the stages adopt the $ConvK$th feature map to assign the proposals and the corresponding predicted boxes, PBRnet is able to leverage the fine-grained information by gradually mapping bounding boxes to low-level feature maps. Secondly, PBRnet extracts local boundary areas of objects for regression which is more precise and more focused on information related to the boundaries. Besides, PBRnet does not involve any process of re-sampling proposals and re-assigning ground truth, and only the top score proposals are selected and transferred for finer refinement. As a result, PBRnet brings less extra computational and memory costs. Furthermore, PBRnet is also independent of the detector architecture and backbones, and can be combined with almost all the state-of-the-art detector to further improve the localization accuracy.

In this work, by using the Faster R-CNN-FPN\cite{lin2017feature,ren2015faster} and ResNet\cite{he2016deep} as backbone, PBRnet brings about 3 point of the $mAP$ gains on COCO2014 benchmark\cite{lin2014microsoft}. It achieves a comparable performance to the Cascade R-CNN\cite{cai2018cascade} but introduces much less extra memory and parameters. Furthermore, by combining the PBRnet with Cascade R-CNN, the $mAP$ is further increased by 1.5, yielding a performance boosting by as high as 5 point.

\section{Methodology}
We developed the PBRnet based on two stage object detection framework, which proposes a multi-stage localization procedure and leverages coarse-to-fine information in different stages of regression to gradually refine the bounding boxes and decreasing the localization errors. FPN\cite{lin2017feature} is  adopted as the backbone.  In the following, the training and testing pipeline of multi-stage regression is elaborated(Section \ref{PBR}), and then boundary area parameterization (Section \ref{param}) and network architecture(Section \ref{architecture}) are detailed.

\subsection{Pyramidal Bounding Box Refinement}\label{PBR}
\begin{figure}
\begin{center}
\includegraphics[width=\textwidth]{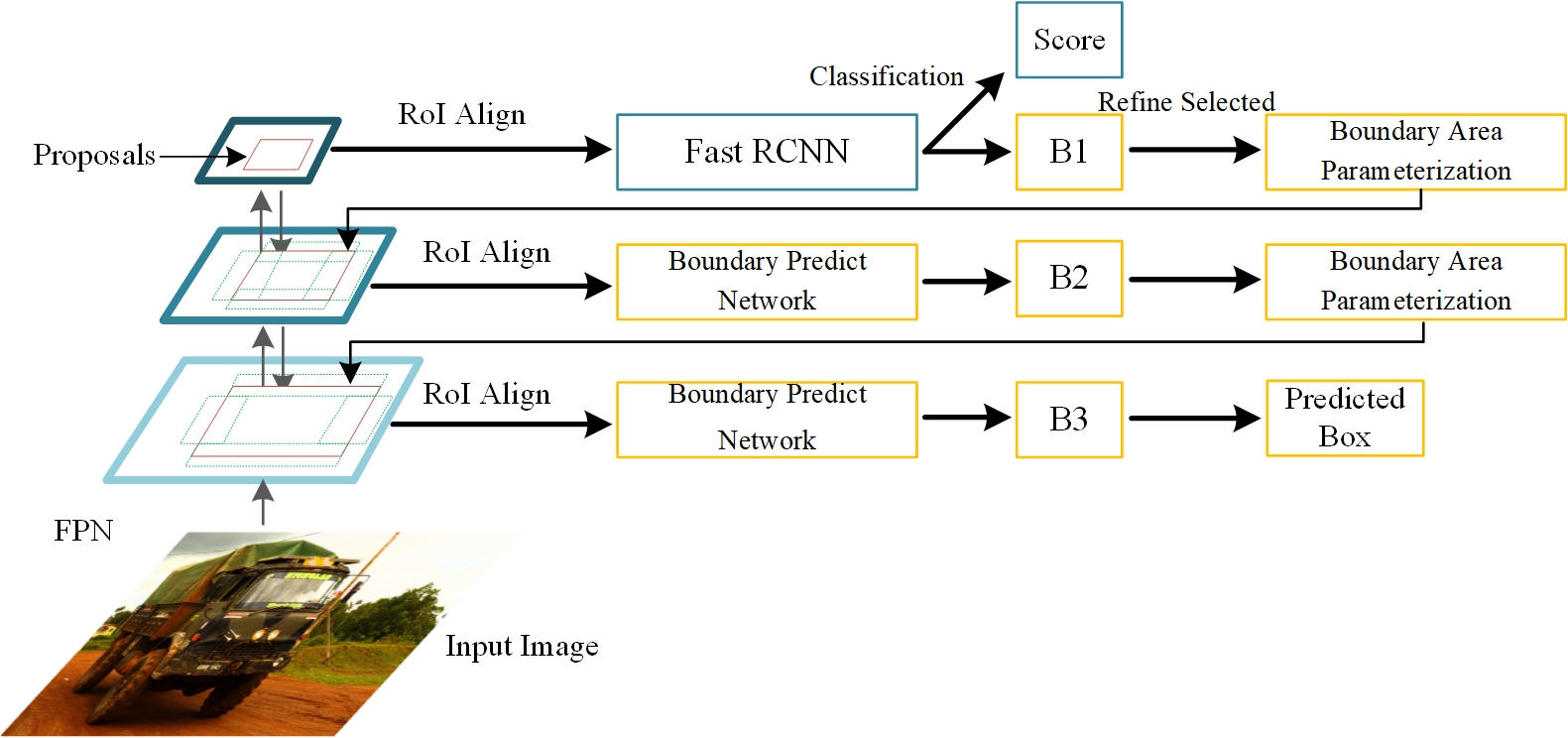}
\end{center}
   \caption{Pipeline of pyramidal bounding box refinement.}
\label{fig3.1}
\end{figure}

Coarse-to-fine framework contains several stages that regress proposals from coarse-grain to fine-grain, and obtains more accurate localization gradually. we proposed a coarse-to-fine architecture named as pyramidal bounding box refinement(PBR) , which extracts gradually fined features in different stages to improve localization accuracy.  The overall pipeline is shown in Figure \ref{fig3.1}. Specifically, the first stage of PBRnet is the same as the localization branch of FPN. We assume a positive proposal $P_0$ produced by RPN which has width $w_0$ and height $h_0$ in the input image, its corresponding ground truth is $G$, which is originally assigned on the $k$th level of feature pyramid $C_k$($k\in{2,3,4,5}$, $C_2$ and $C_5$ indicates the feature map with largest and smallest resolution in FPN respectively), and is then regressed to a predicted bounding box $B_1$. In the next stage, predicted bounding box $B_1$ acts as the proposal in which four boundary areas are extracted as  $P_1^l, P_1^u, P_1^r, P_1^b$. These boundary areas are then pooled on the next level of feature map(remain the same level if it is already on the last feature pyramid $C_2$), in which a boundary predict network is applied to learn the displacement between the boundary of $B_1$ and $G$, and then refine it according to $G$(see Section \ref{architecture}). The boundary refining process can be applied iteratively.

As pointed in \cite{cai2018cascade}, the iterative bounding box regression procedure usually has a central problem that distribution of input in different stages may change significantly as it is more close to the ground truth.  We alleviate this issue by decreasing the extracted boundary areas as the refinement process goes on. The decreased boundary areas can also make the network more focus on the boundary related information in more fine-grained feature maps.

Different from Cascade R-CNN \cite{cai2018cascade} which resamples proposals for more accurate detection, PBRnet is only the coarse-to-fine refinement of initial predicted bounding box and thus do not need to reset ground truth labels and locations for any proposals. The targets of positive proposal $P_t$ in all stages of localization are set as the original assigned ground truth $G$. For boundary refinement stages, only the top score class of each proposal is transferred for refinement when inferencing, and only positive proposals are involved in training. As a result, it adds less extra computational and memory costs.
\begin{figure}
\begin{center}
\includegraphics[width=\textwidth]{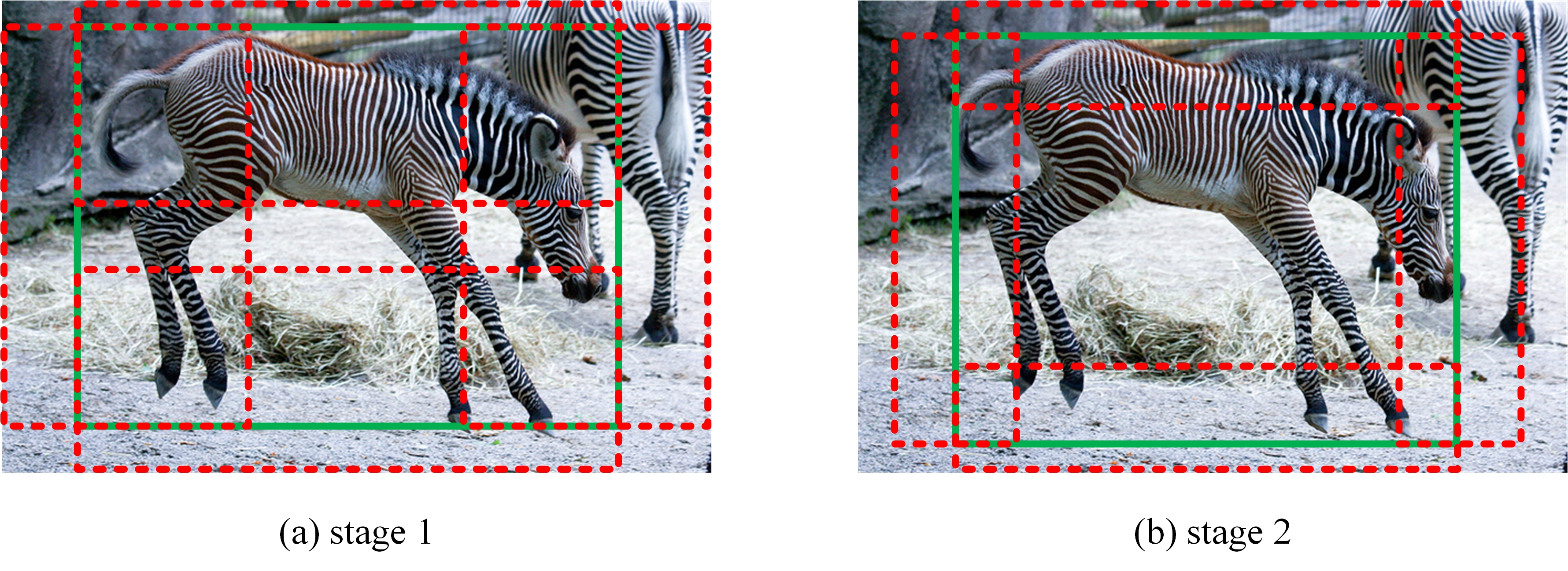}
\end{center}
   \caption{Illustration of four boundary areas extracted in each stage. Green boxes indicate predicted boxes in stage 1 and stage 2.  Red boxes represent four boundary areas taken as input to the next stages with decreasing width/height.}
\label{fig3.3}
\end{figure}

\subsection{Boundary Area Parameterization}\label{param}
Boundary area parameterization takes an essential role in our architecture. For FPN, proposals of different scales are assigned and pooled on feature maps with different resolutions according to its scale. In the $t+1$th($t\ge1,t=1$ indicates the detection process before refinement) stage of boundary refinement, four boundary areas $P_{t}^{l}, P_{t}^{r}, P_{t}^{u}, P_{t}^{b}$  of  box $B_{t}$ predicted in stage $t$,  is taken as input to the boundary predict network(BPN) to predict more precise bounding box  $B_{t+1}$. $P_{t}^{l}$  take the left boundary of $B_{t}$  as center line and expand it to an area with width $c_{t+1} \times w_{t}$ and height $h_{t}$($w_{t}$ and $h_{t}$ are the weight and height of $B_{t}$) , similar procedure is also applied to obtain $P_{t}^{r}, P_{t}^{u}, P_{t}^{b}$ . Assuming $B_{t}$  with coordinates $(x_1^t,y_1^t,x_2^t,y_2^t)$ , the coordinates of $P_{t}^{l}, P_{t}^{r}, P_{t}^{u}, P_{t}^{b}$  are set as:
\begin{equation}
\begin{split}
&P_{t}^{l}=((1+\frac{c_{t+1}}{2})x_1^t-\frac{c_{t+1}}{2}x_2^t,y_1^t,(1-\frac{c_{t+1}}{2})x_1^t+\frac{c_{t+1}}{2}x_2^t,y_2^t)\\
&P_{t}^{r}=(\frac{c_{t+1}}{2}x_1^t+(1-\frac{c_{t+1}}{2})x_2^t,y_1^t,-\frac{c_{t+1}}{2}x_1^t+(1+\frac{c_{t+1}}{2})x_2^t,y_2^t)\\
&P_{t}^{u}=(x_1^t,(1+\frac{c_{t+1}}{2})y_1^t-\frac{c_{t+1}}{2}y_2^t,x_2^t,(1-\frac{c_{t+1}}{2})y_1^t+\frac{c_{t+1}}{2}y_2^t)\\
&P_{t}^{b}=(x_1^t,\frac{c_{t+1}}{2}y_1^t+(1-\frac{c_{t+1}}{2})y_2^t,x_2^t,-\frac{c_{t+1}}{2}y_1^t+(1+\frac{c_{t+1}}{2})y_2^t)\\
\end{split}
\end{equation}

It is worth mention that when the expansion of boundary areas exceeds the shape of input image, for example the left boundary of $P_{t}^{l}$ obtains a negative coordinate,  it will be truncated to zero and the right boundary of $P_{t}^{l}$ will be extended until $P_{t}^{l}$  has the width of $c_{t+1} \times w_{t}$  (In this situation, $P_{t}^{l}=(0,y_1^t,c_{t+1}(x_2^t-x_1^t),y_2^t)$). Same adjustment also applies on  $P_{t}^{r}, P_{t}^{u}, P_{t}^{b}$. Besides, as the localization becomes more and more accurate, we applied a gradually reduced boundary areas in which $c_{t+1}=\frac{1}{2^{t}}$ to produce high quality proposals. The boundary areas of different stages are shown in Figure \ref{fig3.3}. Consequently, $P_{t}$  is assigned to the previous feature map(if have) than $B_{t}$.

The target of refinement is based on the displacement between boundaries of ground truth $G(x_1^*,y_1^*,x_2^*,y_2^*)$  and center line $M_{t}^{l}, M_{t}^{r}, M_{t}^{u}, M_{t}^{b}$ of the corresponding boundary areas. Assume  $\sigma_{t+1}^{*}=(\sigma_{t+1}^{l*},\sigma_{t+1}^{r*},\sigma_{t+1}^{u*},\sigma_{t+1}^{b*})$ and $\sigma_{t+1}=(\sigma_{t+1}^{l},\sigma_{t+1}^{r},\sigma_{t+1}^{u},\sigma_{t+1}^{b})$  are the relative displacement of the four boundaries and its corresponding center of boundary areas:
\begin{equation}
\begin{split}
&\sigma_{t+1}^l=\frac{x_{1}^{t+1}-M_{t}^{l}}{c_{t+1}*(x_2^t-x_1^t)},\sigma_{t+1}^{l^*}=\frac{x_1^*-M_{t}^{l}}{c_{t+1}*(x_2^t-x_1^t)}\\
&\sigma_{t+1}^r=\frac{x_{2}^{t+1}-M_{t}^{r}}{c_{t+1}*(x_2^t-x_1^t)},\sigma_{t+1}^{r^*}=\frac{x_2^*-M_{t}^{r}}{c_{t+1}*(x_2^t-x_1^t)}\\
&\sigma_{t+1}^u=\frac{y_{1}^{t+1}-M_{t}^{u}}{c_{t+1}*(y_2^t-y_1^t)},\sigma_{t+1}^{u^*}=\frac{y_1^*-M_{t}^{u}}{c_{t+1}*(y_2^t-y_1^t)}\\
&\sigma_{t+1}^b=\frac{y_{2}^{t+1}-M_{t}^{b}}{c_{t+1}*(y_2^t-y_1^t)},\sigma_{t+1}^{b^*}=\frac{y_2^*-M_{t}^{b}}{c_{t+1}*(y_2^t-y_1^t)}\\
\end{split}
\end{equation}
The refinement loss is defined as:
\begin{equation}
L_{ref} = \sum_{t=1}^{T}{\sum_{d=l,r,u,b}{Smooth L1(\sigma_{t}^d, \sigma_{t}^{d^*})}}
\end{equation}
As the refinement process applies iteratively, the predicted boundaries on the $t+1$th stage is refined as :
\begin{equation}
\begin{split}
x_1^{t+1}=M_{t}^{l}+c_{t+1}*(x_2^t-x_1^t)*\sigma_{t+1}^l\\
x_2^{t+1}=M_{t}^{r}+c_{t+1}*(x_2^t-x_1^t)*\sigma_{t+1}^r\\
y_1^{t+1}=M_{t}^{u}+c_{t+1}*(y_2^t-y_1^t)*\sigma_{t+1}^u\\
y_2^{t+1}=M_{t}^{b}+c_{t+1}*(y_2^t-y_1^t)*\sigma_{t+1}^b\\
\end{split}
\end{equation}

\begin{figure}
\begin{center}
\includegraphics[width=\textwidth]{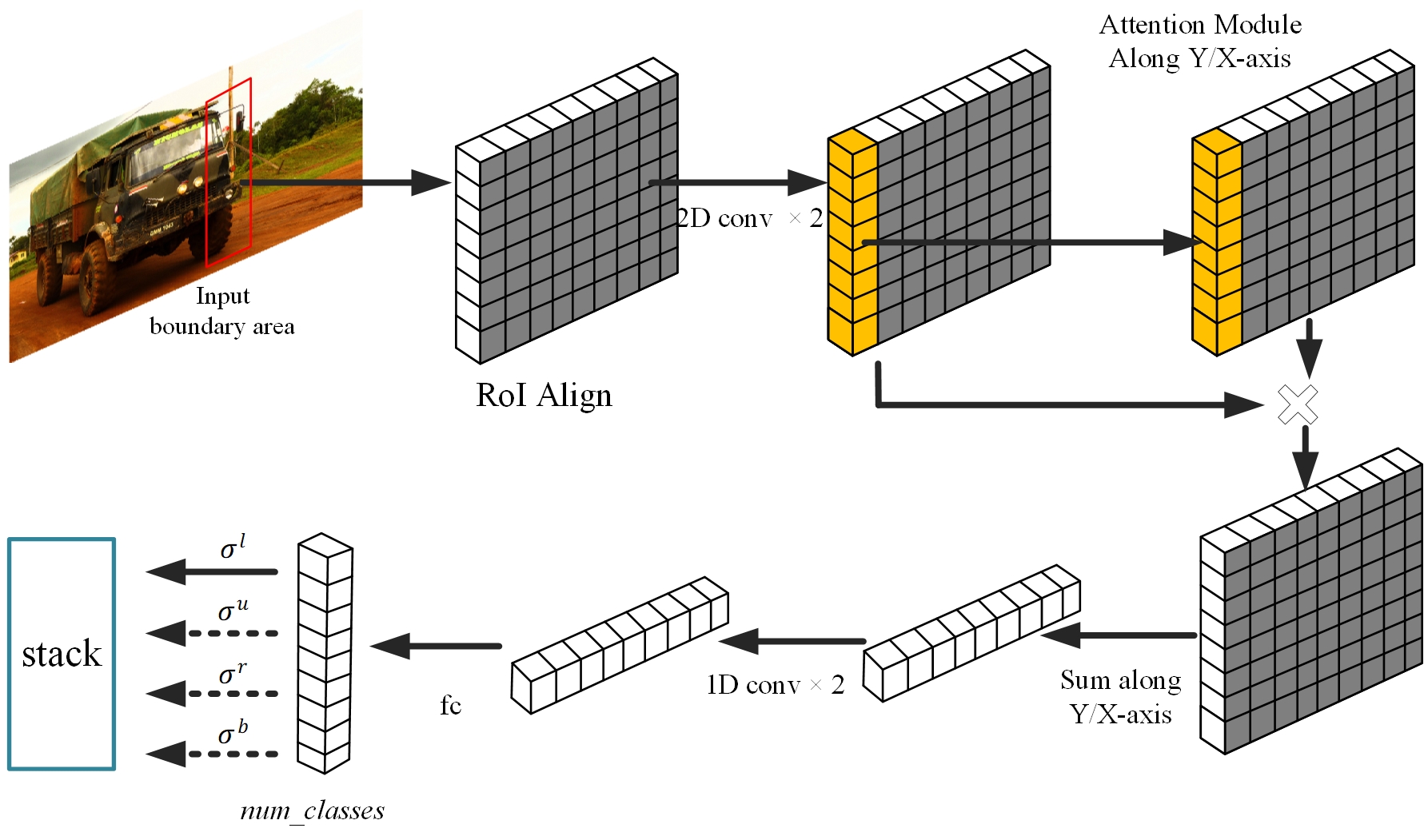}
\end{center}
   \caption{Architecture of the boundary predict network. ``fc'' represents full connected layer, ``conv'' represents convolutional layer.}
\label{fig3.4}
\end{figure}
There can be an issue here that as the multiplier of $\sigma_{t+1}^{*}$  is regulated in different stages by $c_{t+1}$ which decreases rapidly. As a result, the value of $\sigma_{t+1}^{*}$ may increases rapidly and outliers may bring large loss value which can further cause instability of training. To deal with this issue, we applied a truncation operation to keep $\sigma_{t+1}^{*}$ in a range $(-q,q)$ at the beginning of each refinement stage.

\subsection{Boundary Predict Network} \label{architecture}
After the four boundary areas $P_{left}, P_{right}, P_{up}, P_{bottom}$ is obtained, RoI allign layer is performed on them and extract four corresponding boundary descriptions $F_{left}, F_{right}, F_{up}, F_{bottom}$,and are then transferred into four mirrored boundary predict network to predict $\sigma_{t}^{l,r,u,b}$ aforementioned.  The boundary predict network is similar with the side-aware feature extraction network in \cite{wang2019side} but for simplicity we discard the two-step bucketing scheme and bucketing-guided rescoring scheme.

Taken the left boundary predict network as an example. Firstly, a $d\times k\times k$ feature map $F_{left}$ is obtained after ROI Align of the left boundary area(here $d=256$ is the channel number and $k=7$ is the feature map size). It is then passed through two $3\times 3$ convolutional layers, as \cite{cheng2018revisiting,wu2019double} claims that convolutional layer is more capable than fully-connected layer for localization. After that, a self-attention module\cite{vaswani2017attention,xu2015show,hu2018relation,hu2018squeeze}  with group convolution layer and softmax layer along Y-axis is utilized to produce an $1\times k\times k$ attention map, and then boundary features are enhanced by performing weighted sum with the attention map. The new feature map is then summed with the Y-axis(direction parallel to the boundary direction)to obtain a $d\times1\times k$ feature vector. Finally, two $1\times 3$ convolutional layers followed by a fully-connected layer with units of $num\_classes$(indicate the number of predefined object classes) are performed, which the unit related to the object class indicates the prediction of displacement. Finally, the four outputs based on the four boundary areas are stacked as the final refinement location of the predicted bounding boxes.

\section{Experiments}
In this section, we provided model evaluation of PBRnet when applying on several state-of-the-art object detectors, based on credible Microsoft COCO2014 benchmark containing 80 category labels of objects. FPN is adopted as the base detector which all the experiments and models are implemented on. Following conventional practices, we performed training consists of 80k images, validation on the 40k images validation set and performance reporting on the testing set consists of 40k images. Standard COCO metrics are adopted to evaluate our model comprehensively, including $mAP$ (averaged precision over different intersection-over-union thresholds), $AP_{50}$, $AP_{75}$ ($AP$ value when using different IoU thresholds), and $AP_S$, $AP_M$, $AP_L$ ($AP$ at objects of different scales: small,medium,large). Small, medium, large objects are defined as whose sizes are less than 32*32, more than 32*32 and less than 96*96, more than 96*96 respectively.

We firstly provided several ablation studies to determine the architecture and hyper-parameters of the new design and summarized in Section \ref{design}, and then main results are shown in Section \ref{main} to demonstrate the effectiveness of PBRnet.

\subsection{Training and Inference Settings}
All the experiments are implemented on the $mmdetection$ benchmark\cite{chen2019mmdetection}(\footnote{https://github.com/open-mmlab/mmdetection}), and are trained end-to-end on 8 nvidia GTX 1080 GPUs. During training, all models are optimized by synchronized SGD with a weight decay of 0.0001 and momentum of 0.9. Each mini-batch has 2 images, so the effective batch-size is 16. We resize the shorter edge of the image to 800 pixels, and limit the longer edge to less than 1333 pixels to avoid too much memory cost. The typical 1x training settings as in $mmdetection$ which consists 12 epochs is adopted for training. Learning rate is set to be 0.02 at the beginning of the training, and then decreased by a factor of 0.1 on the 9th epoch and 12th epoch. Linear warm-up of learning rate with a ratio of $\frac{1}{3}$ is adopted for the first 500 iterations.

All the models are initialized with FPN weights pre-trained on ImageNet datasets. Following typical practices, the parameters in ResNet are frozen and batch normalization is also fixed during the Fast R-CNN training stage. We only adopted a simple horizontal image flipping data augmentation with flip ratio of 0.5. During training and validation, we pick up 2000 proposals with highest scores on each level of the feature map, and then followed by non maximum suppression (NMS) operation to obtain at most 2000 RoIs for training. During testing, we  pick up 1000 proposals with highest scores for each level of feature map, and then followed by NMS to obtain at most 1000 RoIs. The threshold of NMS is 0.5. RoI-Align is also adopted to obtain the $7\times 7$ feature map for each boundary area.

For PBRnet, unless explicitly stated, $num\_stages$ is set to be 3 which means the refinement process are implemented twice. The $SmoothL1 Loss$ with weight of 0.67 is adopted during the boundary refinement process. To balance the loss terms in different stages, normalization factor is set to be 0.1 for the displacement of central points and 0.2 for the changing of scales in the Fast R-CNN training stage, and are then set as 0.2 and 0.3 for each boundary refinement stage respectively.

\begin{table}
\caption{Performances of PBRnet by applying different truncation values of $q$. }\label{tab5}
\begin{center}
\begin{tabular}{c|c c c c c c }
 \hline
  & mAP& $AP_{50}$ & $AP_{75}$ & $AP_{small}$& $AP_{medium}$& $AP_{large}$\\
 \hline
 No Truncation & 37.2& 55.8& 39.7& 20.4& 40.2& 46.0\\
 -1 to 1 & 37.3& 56.2& \textbf{40.0}& \textbf{20.8}& 40.4& 45.9\\
 -0.5 to 0.5 & \textbf{37.5}& \textbf{56.3}& \textbf{40.0}& 20.7& \textbf{40.6}& \textbf{46.6} \\
 \hline
\end{tabular}
\label{tab5}
\end{center}
\end{table}
\begin{table}
\caption{The comparison of input area by using four boundary areas and the proposal directly. $fc$ layer indicates 2 fully connected layer, the same as the head of FPN. $BPN$ means boundary predict network.}\label{tab3}
\begin{center}
\begin{tabular}{c|c|c|c c c c c c }
 \hline
Refined Object &Refinement Network&mAP& $AP_{50}$ & $AP_{75}$ & $AP_{small}$& $AP_{medium}$& $AP_{large}$\\
 \hline
 proposal & fc layer&36.5& 55.4& 39.3& 19.9& 39.2& 45.6\\
 boundary areas &fc layer &37.2& 56.0& 39.7& 20.5& 40.3& 45.9 \\
 boundary areas &BPN &\textbf{37.5}& \textbf{56.3}& \textbf{40.0}& \textbf{20.7}& \textbf{40.6}& \textbf{46.6} \\
 \hline
\end{tabular}
\label{tab3}
\end{center}
\end{table}

\subsection{Architecture and Hyper-parameters Design} \label{design}
The design of PBRnet is based on several groups of ablation experiments, which is provided as following.  All the experiments are implemented based on FPN with backbone of ResNet50.

\subsubsection{Truncation of $\sigma_{t}^{*}$}  As mentioned in Section\ref{PBR}, along with the boundary refinement goes on, the  displacement $\sigma_{t}^{*}$ may increase rapidly and bring some instability of training. To deal with this issue, we applied a truncation operation on $\sigma_{t}^{*}$ at the beginning of each refinement process to keep the value of it in the range $(-q,q)$. Table \ref{tab5} shows the performance on different values of $q$. We adopted $q$ as 0.5 which brings 0.3 $mAP$ of performance improvement.

\subsubsection{Refinement Method}  Comparison experiments on refinement method is performed and summarized in Table \ref{tab3}, which demonstrates the advantage of using the four boundary areas over using the proposal directly(proposed in Cascade R-CNN\cite{cai2018cascade}) for refinement. By using boundary areas instead of proposals as refined objects but keep the 2 fully connected layer structure(used in Cascade R-CNN\cite{cai2018cascade}), there is an increment of $mAP$ by 0.7.  Replacing the $fc$ layer by boundary predict network also brings 0.3 $mAP$ improvement.

\subsubsection{Number of Stage} The impact of number of stages is summarized in Table \ref{tab2}. The first refinement procedure brings a significant performance boosting by 3.7 point, and the second refinement procedure continues improve the performance slightly.
\begin{table}
\caption{The performance of each stage of refinement. Stage 1 indicates the prediction results before refinement.}\label{tab2}
\begin{center}
\begin{tabular}{c|c c c c c c }
 \hline
 $num\_stages$ & mAP& $AP_{50}$ & $AP_{75}$ & $AP_{small}$& $AP_{medium}$& $AP_{large}$\\
 \hline
 1 & 33.5& 56.0& 35.9& 18.8& 36.5& 41.1\\
 2 & 37.2& \textbf{56.3}& 39.8& \textbf{20.7}& 40.4& 46.1\\
 3 & \textbf{37.5}& \textbf{56.3}& \textbf{40.0}& \textbf{20.7}& \textbf{40.6}& \textbf{46.6} \\
 \hline
\end{tabular}
\label{tab2}
\end{center}
\end{table}

\begin{table}
\caption{Combination and Comparison of PBRnet with state-of-the-art detectors.}\label{tab1}
\begin{center}
\begin{tabular}{c|c |c c c c c c}
 \hline
 &backbone & AP& $AP_{50}$ & $AP_{75}$ & $AP_{small}$& $AP_{medium}$& $AP_{large}$\\
 \hline
 FPN & res50 &34.4 &56.5 &36.7 & 19.4& 37.3& 42.3  \\
 Libra & res50 &35.8 &56.2 &38.8 & 20.4& 38.1& 43.6\\
 Cascade & res50 &37.9 &55.7 &41.1 & 20.1& 40.2& 48.1\\
 PBRnet & res50 & 37.5& 56.3& 40.0& 20.7& 40.6& 46.6 \\
Libra+PBRnet& res50 & 38.7&57.2& 41.3& \textbf{21.8}& 41.2& 47.8\\
Cascade+PBRnet& res50 & \textbf{39.4}& \textbf{57.3}& \textbf{42.1}& 21.3& \textbf{42.1}& \textbf{49.9}\\
 \hline
 FPN & res101 &36.6 &\textbf{58.7} &39.5 & 20.4& 40.0& 46.1\\
 PBRnet & res101 & \textbf{39.4}& 58.3& \textbf{42.2}& \textbf{21.8}& \textbf{42.8}& \textbf{49.6}\\
 \hline
\end{tabular}
\label{tab1}
\end{center}
\end{table}

\subsection{Main Results} \label{main}
\subsubsection{PBRnet on Two-Stage Object Detectors} To show the effectiveness of our model, we firstly applied the PBRnet on the state-of-the-art two-stage object detection models, FPN and Libra R-CNN\cite{pang2019libra}, and performed several groups of comparison experiments. As shown in Table \ref{tab1}, implementing PBRnet on FPN can bring performance boosting by roughly 3 point for either ResNet50 or ResNet101 backbone. PBRnet can also improve the $mAP$ by 2.9 based on the Libra R-CNN baseline. The significant performance gains on various object detection architectures show that PBRnet is an efficient and effective refinement method for predicted bounding boxes. Not surprisingly, since PBRnet mostly focus on fine localization of objects, the improvement of  $AP_{75}$  is much more significant than $AP_{50}$, same as the Cascade R-CNN. Interestingly, the improvement of PBRnet is small on the small objects compared to that on the middle and large objects, this may because that most of small objects are already on the lowest level of feature map and thus benefit little from our coarse-to-fine framework.
 \begin{table}
\caption{Parameters number and Memory costs for FPN, PBRnet and Cascade R-CNN. The initial Parameters of FPN is 25M(ResNet50)+14M(head). (M):$10^6$,(Mb): Megabyte, (s): Second. }\label{tab6}
\begin{center}
\begin{tabular}{c|c |c |c }
 \hline
  & Parameters& Training Speed & Memory During training \\
 \hline
 FPN & 25M+14M & 0.029s& 3793Mb\\
 PBRnet & 25M+21M& 0.040s& 3994Mb\\
 Cascade & 25M+42M& 0.039s& 4112Mb\\
 \hline
\end{tabular}
\label{tab6}
\end{center}
\end{table}
\begin{table}
\caption{ The stage performance of Cascade R-CNN with/without combining to PBRnet. $\overline{1-3}$ indicates the ensemble of three classifiers on the 3rd stage proposals.}\label{tab7}
\begin{center}
\begin{tabular}{c|c |c }
 \hline
  \multirow{2}{*}{Testing stage}& \multicolumn{2}{|c}{ResNet 50}\\
 \cline{2-3}
 & Cascade& Cascade w/PBRnet\\
 \hline
  1& 34.8& 35.3\\
   2& 37.4& 39.0\\
    3& 37.1& 38.5\\
     $\overline{1-3}$& 37.9& 39.4\\
     \hline
\end{tabular}
\label{tab7}
\end{center}
\end{table}

The parameters number and memory cost are shown in Table \ref{tab6}. PBRnet achieves roughly same speed but less extra memory and parameters comparing to Cascade R-CNN .

\subsubsection{Combine PBRnet with Cascade R-CNN} 
\begin{figure}
\begin{center}
\includegraphics[width=\textwidth]{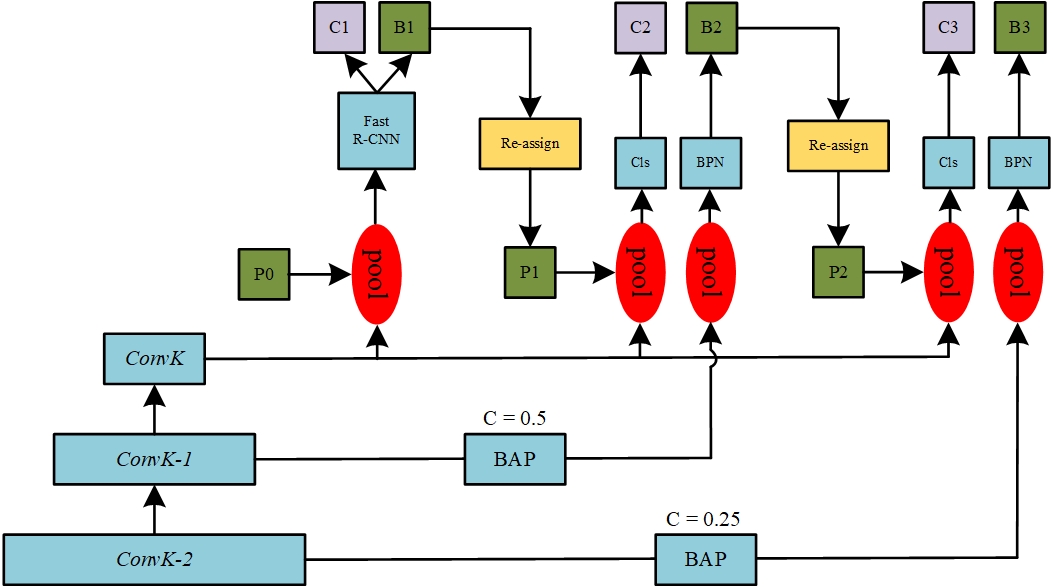}
\end{center}
   \caption{Architecture of applying PBRnet to Cascade R-CNN. ``$ConvK,ConvK-1,ConvK-2$'' represent feature pyramid, ``pool'' region-wise feature extraction, ``Re-Assign'' indicates the process of re-assigning ground truth to proposals with larger IoU threshold, ``P0,P1,P2'' represent input proposals in each stage with gradually improved quality, ``Cls'' classification branch, ``BPN'' boundary predict network, ``BAP'' boundary area parameterization, C1,C2,C3 and B1,B2,B3 indicate classification scores and predicted bounding boxes in each stage.}
\label{fig4.1}
\end{figure}

The performance of PBRnet is further verified by combining it with Cascade R-CNN, which is served as one of the best coarse-to-fine framework. We keep the re-sampling proposals and re-assigning ground truth, as well as the classification process on each stage of Cascade R-CNN, and modify its regression branch by:(1) adopting the boundary areas instead proposals, and applying the boundary predict network for refinement; (2) mapping the bounding box to the lower-level feature map every iteration. The structure is depicted in Figure \ref{fig4.1}. The localization branch of three stages of Cascade R-CNN are replaced by the regressor of PBRnet.  Same as Cascade R-CNN, all regressors are class agnostic. Surprisingly, by simply replacing the regressors in Cascade by that in PBRnet, the new structure surpasses Cascade R-CNN by a significant margin, which leads 1.5$mAP$ of performance boosting (as shown in Table \ref{tab1}), the overall improvement of this new structure can be as high as 5 point comparing to the baseline FPN.

The stage performance is also summarized in Table \ref{tab7}. Interestingly,  the original Cascade R-CNN and Cascade R-CNN+PBRnet shows similar trends of performance improvement in different stages. The 2nd stage improves the performance substantially, and the 3rd stage slightly drops the $mAP$. The ensemble of all classifiers is the best overall.

\subsubsection{Visualization of Results} 
\begin{figure}
\begin{center}
\includegraphics[width=\textwidth]{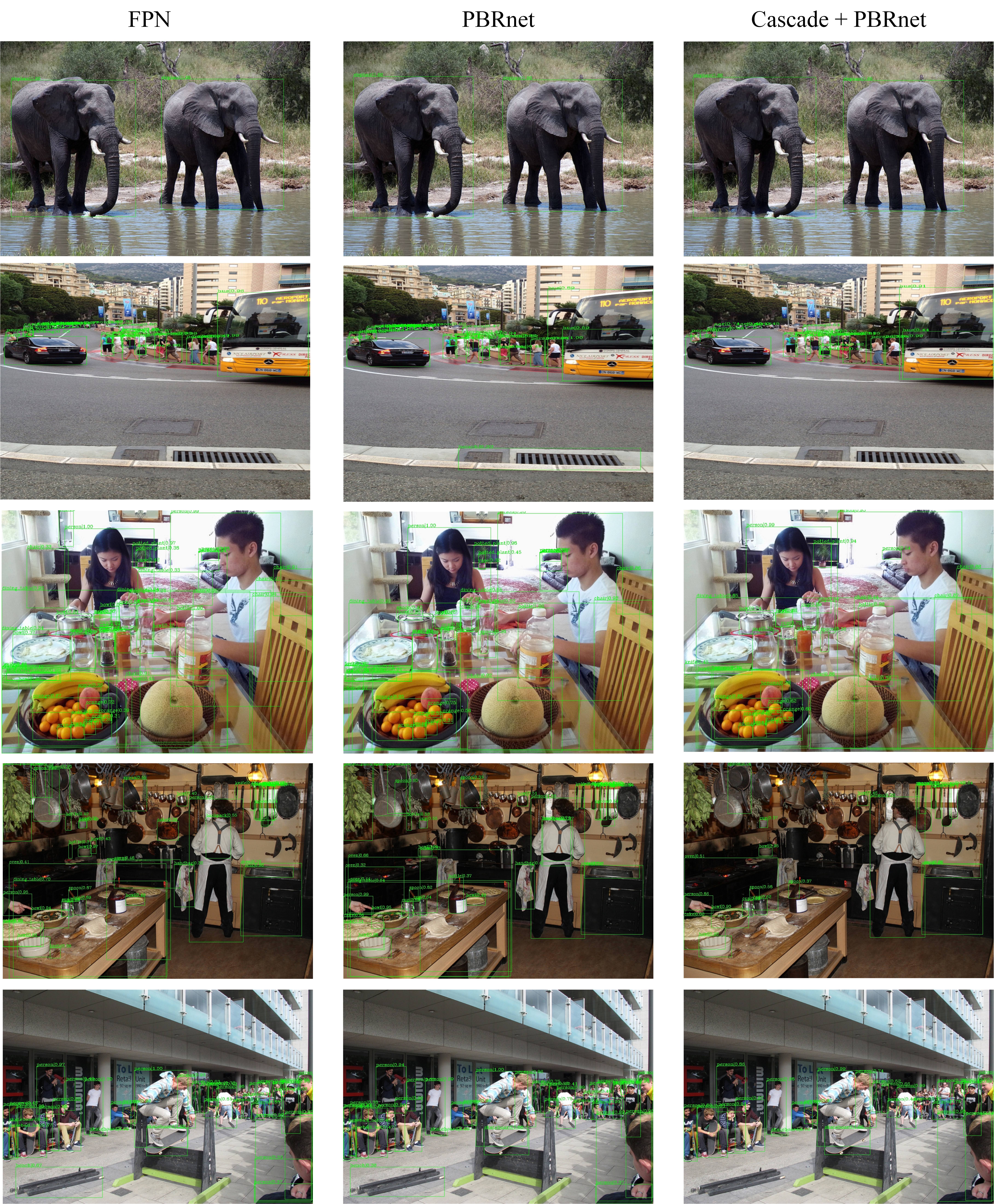}
\end{center}
   \caption{Illustration of bounding box prediction results based on FPN, FPN+PBRnet and Cascade+FPN+PBRnet. All predicted boxes with score higher than 0.3 are shown.}
\label{fig4.2}
\end{figure}

Figure \ref{fig4.2} show some object detection results comparison between FPN, FPN+PBRnet and Cascade+FPN+PBRnet. The results reveal that localizations are more precise by adding the PBRnet, After combining PBRnet with Cascade R-CNN, the classification scores are also refined and improved.

\section{Conclusion}
In this paper, we proposed a novel coarse-to-fine multi-level framework, PBRnet, to improve the object detection performance. The PBRnet refines the predicted bounding boxes iteratively and during each procedure it leverages information of lower level feature maps to achieve a gradually fined refinement.  The PBRnet is firstly verified on FPN and Libra R-CNN, bringing about 3 point improvement of $mAP$. And then combining PBRnet with one of the best single model detector, Cascade R-CNN, the new structure can achieve 5 point performance boosting on $mAP$ comparing to the baseline FPN.  Future work may rely on applying PBRnet on some one-stage or anchor-free detectors, as well as on different computer vision tasks.

%
%
\bibliographystyle{splncs04}
\bibliography{ref}
\end{document}